# Diagnosing Heterogeneous Dynamics for CT Scan Images of Human Brain in Wavelet and MFDFA domain

Sabyasachi Mukhopadhyay[1], Soham Mandal[2*], Nandan K Das[1], Subhadip Dey[3], Asish Mitra[4], Nirmalya Ghosh[1], Prasanta K Panigrahi[1]

[1]Department of Physical Sciences, IISER, Kolkata
[2]Department of Computer Science Engineering, IEM, Kolkata
[3]Department of Agricultural Engineering, BCKV, Kalyani
[4]Department of Basic Science & Humanities, CEM, Kolaghat

**ABSTRACT**

CT scan images of human brain of a particular patient in different cross sections are taken, on which wavelet transform and multi-fractal analysis are applied. The vertical and horizontal unfolding of images are done before analyzing these images. A systematic investigation of de-noised CT scan images of human brain in different cross-sections are carried out through wavelet normalized energy and wavelet semi-log plots, which clearly points out the mismatch between results of vertical and horizontal unfolding. The mismatch of results confirms the heterogeneity in spatial domain. Using the multi-fractal de-trended fluctuation analysis (MFDFA), the mismatch between the values of Hurst exponent and width of singularity spectrum by vertical and horizontal unfolding confirms the same.

## 1. INTRODUCTION

Significant research works are being undertaken for understanding of the dynamics of biomedical images over the last few decades. There are several conventional methods for early detection and diagnosis of abnormalities in cell tissues. For quantification of both the morphological and biochemical alterations associated with abnormalities development like cancer, the biomedical imaging based approaches have shown early promise [1]. Wavelets and MFDFA are very useful tool for their versatile applications from the field of environmental science to biomedical imaging [2, 5-8]. In previous works by Mukhopadhyay et.al., the biomedical image analysis of DIC stromal as well as epithelium regions are done with wavelet and MFDFA [7-8]. The significant variations are observed among different grades of cancer tissues for stromal and epithelium regions. In this paper, the results are discussed elaborately about heterogeneous dynamics study of biomedical images.

## 2. THEORY

**2.1 Wavelet Transform:** If the low pass and high pass coefficients are $c_k's$ and $d_{j,k}'s$ respectively, the data set can be expressed for Discrete wavelet transform as $f(t) = \sum_k c_k \phi_k + \sum_k \sum_{j=0}^{\infty} d_{j,k} \psi_{j,k}$. The trend components are extracted by father wavelet $\phi_k$, located at $k$ with level $j$ and the deviations from the trend are picked up by mother wavelets $\psi_{j,k}$.

**2.2 Multi-Fractal De-trended Fluctuation Analysis (MFDFA):** Briefly, the profile $Y(i)$ (spatial series of length N, $i = 1 \ldots N$) is first generated from the one dimensional spatial index fluctuations. The local trend of the series ($y_b(i)$) is determined for each segment $b$ by least square polynomial fitting, and then subtracted from the segmented profiles to yield the de-trended fluctuations. The resulting variance of the de-trended fluctuation is determined for each segment as

$$F^2(b,s) = \frac{1}{s}\sum_{i=1}^{s}[Y\{(b-1)s + i\} - y_b(i)]^2 \qquad (1)$$

*somban637@gmail.com, phone: +91-9433901722

The moment (q) dependent fluctuation function is then extracted by averaging over all the segments as

$$F_q(s) = \left\{ \frac{1}{2N_s} \sum_{b=1}^{2N_s} [F^2(b,s)]^{\frac{q}{2}} \right\}^{1/q} \qquad (2)$$

The scaling behaviour is subsequently determined by analysing the variations of $F_q(s)$ vs s for each values of q, assuming the general scaling function as

$$F_q(s) \sim s^{h(q)} \qquad (3)$$

Here, the generalized Hurst exponent h(q).

## 3. RESULTS AND DISCUSSIONS

The CT scan images of human brain in different cross sections are taken for analysis purpose. The sample CT scan image is mentioned below as an example.

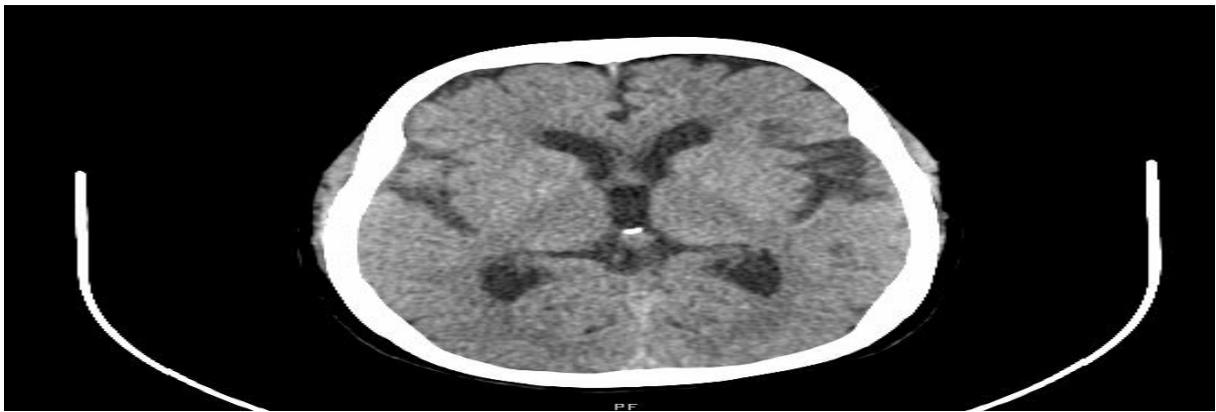

Fig1. CT scan image of human brain

We know that the biomedical images can be considered as irregular signals. Wavelets are ideal analysis tool for irregular data processing purpose. For any observed signal x(t)=f(t)+e(t), where f(t) is the signal and e(t) is the noise, using wavelets f(t) can easily be extracted out. In this paper, Discrete wavelet transform (DWT) through Daubechies basis analysis has been done up to level-5 for identifying localized fluctuations (high pass coefficients) over polynomial trends for clear characterization and differentiation of CT scan images of human brain in different cross sections.

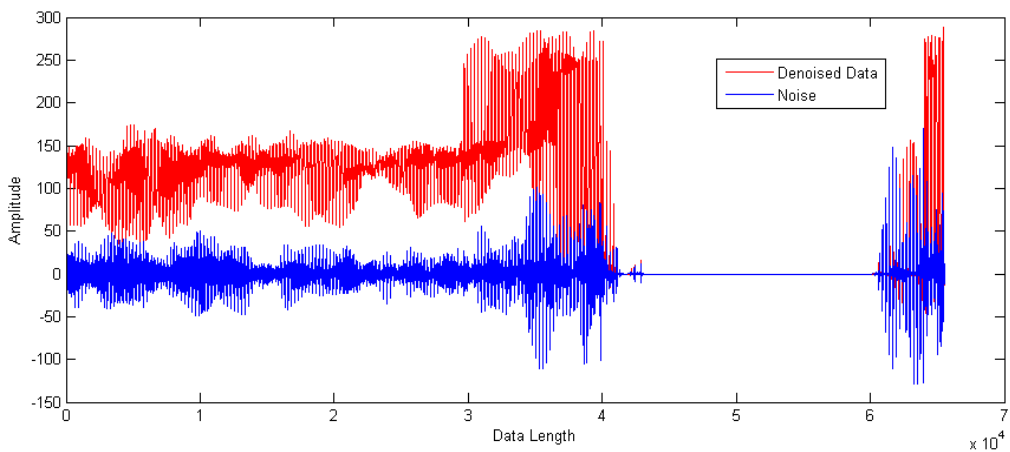

Fig2. Denoised data and Noise (Fluctuations) after applying Daubechies-4 (Db-4) on CT scan images

The plot of Wavelet normalized energy vs level of decomposition (level1 to level5) of high pass coifficients (fluctuations) have been performed here using db-4 for CT scan images of human brain.

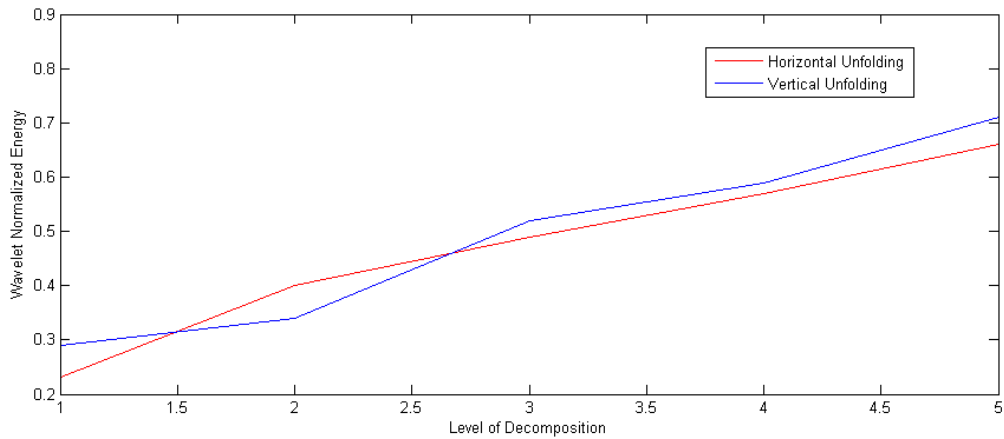

Fig3. Wavelet Normalized Energy for Vertical and Horizontal unfolding

It is clearly observed that the normalized energy value differs from horizontal to vertical unfolding which is the evidence of medium heterogeneity.

Now the semi log plots for horizontal and vertical unfolding components are done to check the anisotropic nature of fluctuations using Morlet transform.

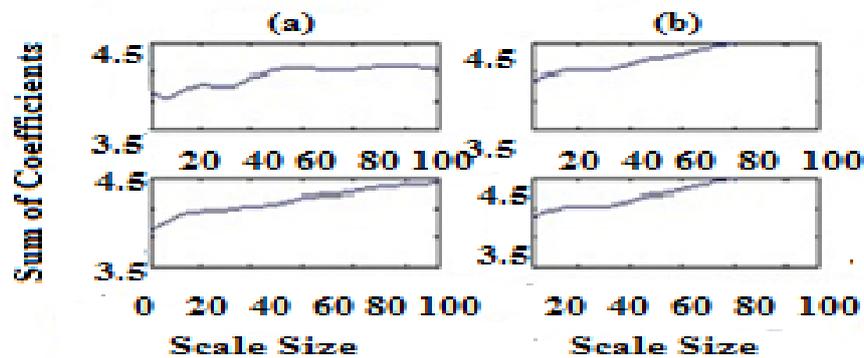

Fig4.Wavelet semi-log plot of brain CT scan images of human brain in different cross sections where column (a) stands for horizontal unfolding and column (b) stands for vertical unfolding.The mismatch among horizontal and vertical unfolding results clearly confirms the heterogeneity

The mismatch between the horizontal and vertical unfolding from fig3 result have confirmed the medium anisotropic nature. It is actually an evidence of medium heterogenity of CT scan images.
We now turn to check the self-similar purpose. In this regard, MFDFA is helpful for subtle morphological changes which are otherwise hidden.

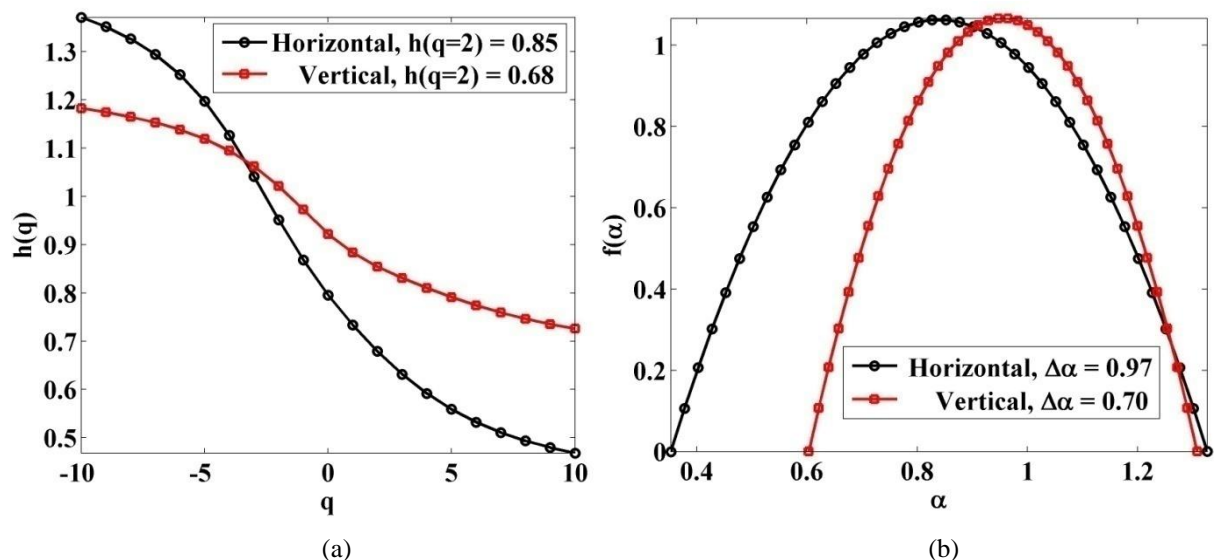
(a) (b)
Fig5 (a) & (b). Plot of Hurst Exponent (Hq) and singularity spectrum [f(α)] for CT scan Images of vertical and horizontal unfolding respectively

The mismatch of hurst exponent and singularity spectrum width values for horizontal and vertical unfolding cases clearly have shown the heterogeneity nature of the human brain CT scan images.

## CONCLUSION

We have described a robust method to automatically detect the heterogeneity in CT scan images of human brain. The current paper emphasizes upon the generalization of the method i.e., selection of wavelets, the scale of MFDFA analysis depend upon the judicious choices of the user. Thus the success of the methods depends upon that judicious choice. Wavelet normalized energy, Wavelet semi-log plots as well as MFDFA based hurst exponent and singularity spectrum based results further depict the heterogeneity of the CT scan images of human brain. In a nutshell, the proposed method is aimed to precisely extract out the substantial features which was unknown using traditional statistical operations over medical images. The wavelet and MFDFA based approach in this paper can aid to overcome those limitations.  As a plan for future work, we are collecting more data for analysis purpose and hope to extract out more prominent features of heterogeneity. It will even help to predict if there is any kind of abnormality occurs in tissues due to affected by chronic diseases. Authors hope that the current study of the heterogeneous dynamics of CT scan images of human brain will help the researchers move this field forward.

## Acknowledgement
The authors thank Bankura Sammilani Medical College and Hospital, Bankura, West Bengal for providing the CT images of human brain in different cross-section.